\documentclass[10pt,onecolumn]{article}
\usepackage{amsmath}
\usepackage{graphicx}
\graphicspath{ {images/} }
\usepackage{lipsum,multicol}
\usepackage{kantlipsum}
\usepackage{caption,subcaption}
\usepackage{pdfpages}
\usepackage{url}
\usepackage{hyperref}
\usepackage{doi}
\usepackage{comment}
\begin{document}

	\title{Contour Integration using Graph-Cut and Non-Classical Receptive Field}
		
		\author{Zahra~Mousavi~Kouzehkanan \\ Reshad~Hosseini\\ Babak~Nadjar~Araabi\\
		\footnotesize \texttt{\{z\_mousavi, reshad.hosseini, araabi\}@ut.ac.ir} \\ }
		\date{}
	
	\maketitle
	\begin{abstract}

		Many edge and contour detection algorithms give a soft-value as an output and the final binary map is commonly obtained by applying an optimal threshold. In this paper, we propose a novel method to detect image contours from the extracted edge segments of other algorithms. Our method is based on an undirected graphical model with the edge segments set as the vertices. The proposed energy functions are inspired by the surround modulation in the primary visual cortex that help suppressing texture noise. Our algorithm can improve extracting the binary map, because it considers other important factors such as connectivity, smoothness, and length of the contour beside the soft-values. Our quantitative and qualitative experimental results show the efficacy of the proposed method.

	\end{abstract}
{\bf Keywords:} Contour integration, Non-classical receptive field, Graph-cut, Surround modulation .
	    \section{Introduction}
	    Contour integration is an important and challenging problem in machine vision. Image contours and edges play a significant role in many tasks such as segmentation and object detection. Alterations in brightness, color and texture  results in production of  myriad  edges in an image. Yet, contours are more general concepts and construct the image boundaries and general form of the objects. From bottom-up view point, there are  two steps to be taken into account for contour integration. The primary step is computing edge map in which the value of a pixel is usually soft and determines how likely it belongs to an object contour. Several methods are developed to calculate the edge map. These methods can be low level such as traditional edge detectors or high level like GPB  \cite{arbelaez2011contour}, SCG  \cite{xiaofeng2012discriminatively}. The second step is converting the edge map to contour map. It means that object contours should be popped out and texture noises should be suppressed.
	    
	    Obviously, the performance of the second step strongly depends on the quality of the edge map. A low level edge map cannot produce a very proper contour map. However, the second step can play a significant role in contour integration process. The most straightforward way is to apply an optimal threshold. Under the threshold, the edges are eliminated and the remaining edges make contours. Some contour integration methods such as Canny \cite{canny1986computational} , GPB \cite{arbelaez2011contour}, SCG \cite{xiaofeng2012discriminatively} do non-maximum suppression before the threshold being applied. Many algorithms have addressed this issue which will be reviewed in Section \ref{chp.related}. A wide group of approaches exercise contextual influences to ameliorate edge map and create contour map  \cite{grigorescu2003contour,petkov2003suppression,grigorescu2004contour,papari2007biologically,zeng2011center,wei2013contour,zeng2011contour,li1998neural,sang2017contour,su2017contour,guy1992perceptual}.

	    
	    Several psychophysical and neurophysiological evidence revealed that contextual influences have an essential role to play in human contour integration process  \cite{hess2014contour,kuai2017contour,qiu2016responses}. Cells in primary visual cortex (V1) are responsive to a narrow range of orientations (approximately 40) within their receptive field  \cite{hubel1962receptive,hubel1968receptive}. Each cell has a special receptive field and a preferred orientation. Therefore, V1 cells all together act as a filter bank and extract oriented line segments from input image. Extracted local line segments should be integrated somehow and form global contours.

	     
	    The sensitivity of each cell in V1 is not limited to its classical receptive field and the surrounding stimuli can modulate it  \cite{hess2003contour,kapadia2000spatial}. This surrounding area is known as non-classical receptive field. The recording from V1 neurons in monkeys indicated that putting a collinear line segment outside the classical receptive field increases the neuronal response and this enhancement decreases by distance and misalignment  \cite{kapadia1995improvement}. Similarly, putting random line segments surrounding the classical receptive filed, often reduces the neuronal response  \cite{kapadia1995improvement}. The analysis of human fMRI responses revealed similar results in early visual areas  \cite{altmann2003perceptual} .

	    The framework we designed is inspired by the concept of non classical receptive field and is able to convert any arbitrary edge maps to a binary contour map. In this framework, beside the edge strengths, we also consider other significant factors such as connectivity, smoothness, and length of the contour. We use conditional random field which is an undirected graphical model and contours were detected by solving a maximum a posteriori problem and assigning a binary label to each edge segment. The probability of each labeling assignment is defined as a Gibbs distribution and consists of unary and pairwise parts that are respectively defined on graph nodes and graph arcs. The Matlab implementation is available on the Github website\footnote{http://github.com/z-mousavi/ContourGraphCut}. Some fundamental characteristics of our framework are: 
	    
	    
	     \begin{enumerate}
	     \item We define energy functions in a way that a maximum posteriori problem is reduced to a min-cut problem in the graph theory. Thus, we can compute optimal MAP estimation far more rapidly. Graph-cut is considered to be a useful tool for energy minimization in machine vision  \cite{boykov2001interactive,boykov2004experimental}.
	     
	     \item In order to define energy functions, the concept of Association Field  \cite{field1993contour} is adopted. Association field determines the edge segment interactions to be facilitatory or inhibitory. The Interaction between two edge segments relies upon their relative position and direction. Association field assists us to consider contextual influence (non-classical receptive field or surround modulation) in our contour detection process.
	     \item In order to train parameters, instead of using common cost functions for conditional random field such as pseudo likelihood, we directly optimize the output of framework. For this purpose, we use global MCS  \cite{huyer1999global} optimization. Due to the swift nature of  MAP inference in our framework, this approach is feasible. Applying this idea improves the framework performance.
	     \end{enumerate}

	   We use BSDS500 dataset  \cite{arbelaez2011contour} to train the parameters of our framework. We observed that our proposed method for producing binary map from soft edge values improves the performance over optimal threshold for different edge-segment extraction methods from low-level to high-level. This paper is organized as follows: in Section~\ref{chp.related} we discuss the related works and the details of our framework are explained in Section~\ref{chp.method}. Finally, we evaluate our framework quantitatively and qualitatively in Section~\ref{chp.result}.
	   

	    \section{Related works}
	    \label{chp.related}
	    In this section we review contour integration methods. These methods commonly aim at popping out contours from a pool of edge segments. Edge segments (depending on the method) are extracted in a specific way for example through using a traditional edge detector. Gestalt psychologists made the first attempts for contour integration. They carried out a number of experiments and figured out  that human visual grouping is influenced by a variety of  factors such as continuity, smoothness, closure, etc.  \cite{wertheimer1938laws}. Contour integration methods fall  into two groups of  bio-inspired and others.

	   \subsection{Bio-inspired methods}

		Contextual influences are of utmost significance in the human contour integration process  \cite{hess2014contour,kuai2017contour,qiu2016responses}. Several psychophysical and neurophysiological evidence demonstrate that the response of human visual system to an oriented stimulus is modulated by surrounding stimuli  \cite{hess2003contour,kapadia2000spatial}. Inspired by this evidence, many contour integration methods have been developed. These methods take the advantage of  surround modulation in order to highlight main contours and suppress texture noise. In the literature, the non-classical receptive field is simulated in various ways.

		Usually, \textit{nCRF} (neural Conditional Random Field) is considered as a ring around the classical receptive field. In papers  \cite{grigorescu2003contour,petkov2003suppression,grigorescu2004contour} solely the surrounding inhibition is considered. The main problem of these methods is contour self-inhibition which leads to suppression in region boundaries. Several solutions have been presented to avoid self-inhibition. Papari et al.  \cite{papari2007biologically} split the nCRF into two truncated half-rings oriented along the edge and only consider the half-ring with minimum inhibition.

		The adaptive nCRF is defined by the latest approaches in a way that the nCRF is split into sub-regions and  diverse modulation effects for each region are defined. A butterfly-formed surrounding area is normally employed so that the side and end sub-regions work in different manners. In  \cite{zeng2011contour}, the end sub-regions have been excluded and the side sub-region with smaller inhibition strength is the only contributor to the final neuronal response. Zeng et al. \cite{zeng2011center} and Wei et al. \cite{wei2013contour} respectively define adaptive inhibitory and disinhibitory manners for end regions. Some approaches consider facilitatory effects for end regions  \cite{li1998neural,sang2017contour,su2017contour}. Facilitatory modulations avoid to suppress region boundaries.
		 
		A number of methods solely implement facilitatory modulations. For example, Guy and Medioni  \cite{guy1992perceptual} developed a facilitation-based voting approach. For this purpose, they used a voting operator called extension field and created a saliency map. An Extension Field describes the association of a single edge segment to its environment in terms of length and direction \cite{guy1992perceptual}. Our proposed method considers both inhibitory and facilitatory effects.

		\subsection{Other methods}
		
	Some contour integration methods adopt voting approaches and try to detect specific structures including lines, circles and smooth curves. Hough transform \cite{hough1962method} is one of the parametric voting methods introduced in this regard. Akinlar and Topal developed a parameter free framework for line and circle detection  \cite{akinlar2011edlines,akinlar2013edcircles}. Geisler et al.  \cite{geisler2001edge} utilized natural image statistics to develop an edge grouping framework based on Bayesian decision theory. The aforementioned edge grouping depends on the position and direction of the edge segments. Cao et al. \cite{cao2019application} employ both normalized difference of Gaussian function and a sigmoid activated function to define an inhibition term and diminish  the background noise.
	
	
	Relaxation labeling is a graph-based framework similar to our proposed method \cite{parent1989trace,hummel1983foundations}. They consider a graph the nodes of which are edge segments and arcs link the surrounding edges. In this method, a labeling vector $P_{N\times1}$  is defined in which  $N$  is the number of edge segments, and  $P(n)$ is a real or binary value that determines the possibility of the nth edge segment belonging to object contours. Suppose $i$ and $j$ are two neighbor edge segments, the function $C(i,j)$ computes the degree of consistency to which object contours both $i$ and $j$ belong and seeks to consider contextual effects. For a specific $P$, an average consistency is computed as $a(P)=\sum_{\substack{i,j}}P(i)P(j)C(i,j)$. The vector $P$ which maximizes $a(P)$ is the final saliency map. The major disadvantage of relaxation labeling is that the suggested solutions are usually iterative and do not guarantee convergence to the global optimum. However, our framework manages to obtain the global optimum swiftly.


		A group of methods called edge linking are aimed at enhancing the binary output of traditional edge detectors through removing the noisy edges and filling edge pixel gaps. In this regard, many and various tools such as graph sequential search  \cite{farag1995edge,ji2013sequential} and ant colony optimization  \cite{lu2008edge} have been employed. Topal and Akinlar \cite{topal2012edge} proposed an edge drawing algorithm. In this algorithm, first, some anchor points are extracted from the image. Then, these anchors are connected via smart routing. In  \cite{akinlar2015cannysr}, the authors suggested CannySR edge detection by applying smart routing algorithm on Canny binary output. A predictive edge linking algorithm has been developed by Akinlar and Chome  \cite{akinlar2016pel}. This method walk over the binary edge map based on previous movement predictions. Seo  \cite{seo2019subpixel} after localizing line segments, adopt a line linking method through which the linking distance of the two lines is calculated as the product of geometric and angular distances. The current line will then be connected to a  line with minimum linking distance.


		\section{Method}
		\label{chp.method}
		
		We propose to formulate contour integration as a labeling problem. We have plenty of edge segments in different parts of an image. Each edge segment has a direction, position and descriptor. The descriptor can be computed through any methods and determines the strength of the edge segment. For example, the edge segments can be pixels with large gradient magnitude; the direction of the edge segment is the direction perpendicular to the gradient direction; and the descriptor can be the gradient magnitude of each edge segment.
		
		In order for us to assign a binary label to each edge segment, we should take the following information into account: If the edge segment is part of a strong contour, it takes label 1. The edge segment takes label 0, if it is isolated or is a part of a weak contour. The strength or weakness of a contour relies on several factors including: the length of contour, the smoothness and strength of constituent segments. Regarding these, we aim at designing a framework for contour integration that considers all of these factors mentioned.
		

		The conditional random field is utilized for solving the labeling problem. Suppose the image $I$ has $M$ edge segments that are represented by $ C(I)=\{f_i,x_{i},y_{i},o_{i}\}_{i=1}^{M} $. $f_i $ and $ o_{i}\in{[0,\pi)}$ are the descriptor and orientation of the edge segment $i$, respectively. $x_{i}$ and $y_{i}$ indicate the position of $i$  in the image plane. In fact, $C(I)$ contains observed variables.
		
		
	   The set ${\{X\} _{i=1}^{M}}$ are random output variables in which $X_{i}\in \{0,1\}$ denotes the binary label of the edge segment i. Markov blanket $N_i$ almost plays the role of non-classical receptive field and determines surround modulation area. Consider a graph in which the set $\{X\} _{i=1}^{M}$ are nodes and arcs are defined between neighboring edge segments based on neighborhood N. The probability of each labeling configuration $\{x_{1}, x_{2}, ..., x_{M}\}$, with respect to $C(I)$, is defined as follows:
	   
	    \begin{subequations}
	    	\allowdisplaybreaks
	    	\begin{align}
	    	p(x_{1}, x_{2}, ..., x_{n} | C(I)) &=\frac{1}{Z(C(I))}{e}^{-E(x_{1}, x_{2}, ..., x_{n} |C(I))} , \label{eq7}  \\
	    	\begin{split}
	    	E(x_{1}, x_{2}, ..., x_{n}|C(I)) &= \sum_{i}{\epsilon_{i}(x_{i}|C(I))} \\
	    	&+ \sum_{i}{\sum_{j\in{N_{i}}}{\epsilon_{i,j}(x_{i},x_{j}|C(I))}}, \label{eq8}
	    	\end{split}
	    	 \\
	    	Z(C(I)) &=  \sum_{x\in{X}}{{e}^{-E(x_{1}, x_{2}, ..., x_{n} |C(I))}}. \label{eq9}
	    	\end{align}
	    \end{subequations}

	The energy or cost of each labeling assignment is computed by equation \eqref{eq8} which consists of two terms. First one is the summation of the unary energies, and the second is the sum over all pairwise energies. In equation \eqref{eq9}, $Z(C(I))$, called partition function, is practically incomputable because it is the sum over all $2^{n}$ labeling configurations.
	
	  \subsection{Association Field}
	  Before specifying unary and pairwise energies, we need to define the concept of association field in a mathematic form. As we discussed earlier, the association field determines the interaction between two neighboring edge segments. We define two functions of $J(i,j)$ and $W(i,j)$ in a way that they get the position and orientation of the edge segments $i$ and $j$, and return the amount of excitation and inhibition, respectively. As an example, if edge segments $i$ and $j$ can construct a smooth contour, the excitation level is high and the inhibition level is low. On a par with paper  \cite{ernst2012optimality}, we use multiplication of the two functions:
	  
	  \begin{subequations}
	  	\begin{align}
	  	J(i,j)&=A_{\text{exc}}(i,j)A^d(i,j),\\
	  	W(i,j)&=A_{\text{inh}}(i,j)A^d(i,j).
	  	\end{align} \label{eq15}
	  \end{subequations}
  
  Functions $A_{\text{exc}}$ and $A_{\text{inh}}$ are angular parts that are defined in equation \eqref{eq16}. In this equation, $o_{i,j}$ is the orientation of a line that connects two edge segments $i$ and $j$, and $a_{1}$ and $a_{2}$ are parameters which should be learned from training images.
    \begin{subequations}
      \begin{align}
          A_{\text{exc}}(i,j)
          =e^{a_{1}\min(|\cos(o_i-o_{i,j})|,|\cos(o_j-o_{i,j}|)+ a_{2}|\cos(o_i-o_j)|}
      \\
      A_{\text{inh}}(i,j)
      =e^{a_{1}\max(|\sin(o_i-o_{i,j})|,|\sin(o_j-o_{i,j}|)+ a_{2}|\sin(o_i-o_j)|}
      \end{align} \label{eq16}
    \end{subequations}

Function $A^d$ in equation \eqref{eq15} conveys the distance factor that is defined in equation \eqref{dis_fun}. Notation $d_{i,j}$ in this equation indicates Euclidean distance between edge segments $i$ and $j$. Undoubtedly, the growth of distance reduces the amount of excitation and inhibition.
	
	 \begin{equation}
	 A^d(i,j)=e^{-\frac{d_{i,j}}{\sigma}}  \label{dis_fun}
	 \end{equation} 
	    
	\subsection{Unary energy}
	Function $\epsilon_{i}(x_{i}|C(I))$ computes energy when edge segment $i$ takes the label $x_{i}$. We, firstly, define probability $p_{i}(x_{i}|C(I))$ as a convex combination of two logistic function:
	
	\begin{subequations}
	\begin{align}
    \begin{split}
	p_{i}(0|C(I)) &=\alpha{\frac{1}{1+e^{(\lambda{f_i}+\omega_{0})}}}\\
	&+(1-\alpha){\frac{1}{1+e^{(\omega_{1}\text{Exi}_i(I))-\omega_{2}\text{Inh}_i(I))}}}
    \label{eq10}
    \end{split}	\\
    p_{i}(1|C(I)) &= 1-p_{i}(0|C(I));\label{eq11}
    \end{align} \label{unary_prob}
	\end{subequations}

Unary energy is defined as $\epsilon_{i}(x_{i}|C(I))=-\log(P_{i}(x_{i}|C(I)))$. Undoubtedly the parameters ${\alpha\in{[0,1]},\ \lambda,\ \omega_0,\   \omega_1\geq0}$ and ${\omega_2\geq0}$ in equation \eqref{unary_prob} should be learned properly. The first logistic term in equation \eqref{eq10} is related to the strength of the edge segment. A strong edge segment probably belongs to a strong contour. Thus, taking label 0 will be costly. Second logistic in equation \eqref{eq10} computes the probability of belonging to a smooth contour. Functions $\text{Exi}_i(I)$ and $\text{Inh}_i(I)$ are defined in equation \eqref{eq12} and equation \eqref{eq13} and determine how much edge segment $i$ is excited and inhibited by its surrounding area, respectively. In equation \eqref{eq12} and equation \eqref{eq13}, $J(i,j)$ and $W(i,j)$ are regulated by $f_j$ . Therefore, strong edge segments have more effects. 
	\begin{subequations}
		\begin{align}
		\text{Exi}_i(I) &= \sum_{j\in{N_i}}{f_jJ(i,j)} \label{eq12} \\
		\text{Inh}_i(I) &= \sum_{j\in{N_i}}{f_jW(i,j)} \label{eq13}
		\end{align}
	\end{subequations}
    
     
     If edge segment $i$ belongs to a smooth contour, the second logistic in equation \eqref{eq10} gets a big value and the cost of getting label 0 increases. In order to understand the importance of the second logistic, consider we have no edge descriptor and all edge segments have equal strength. In this case, first logistic is automatically eliminated and only the second term is effective. Parameter $\alpha$  plays a significant role in unary energy . For example, when we use a very powerful and high level descriptor, $\alpha$  should be more than $0.5$.
   
	\subsection{Pairwise energy}
	
	Function $\epsilon_{i,j}(x_{i},x_{j}|C(I))$ in equation \eqref{eq8} is the pairwise energy that computes the amount of energy when edge segments $i$ and $j$ take labels $x_{i}$ and $x_{j}$, respectively. We define pairwise energy in the format of  Ising model \cite{koller2009probabilistic}:
	\begin{equation}
	\epsilon_{i,j}(x_{i},x_{j}|C(I)) =\\
	\begin{cases}
	0    & \  \text{if } x_{i}=x_{j} \\
	\beta.\mu(i,j)   & \  \text{if }  x_{i}\neq{x_{j}}
	\end{cases} \label{eq14}
	\end{equation}
	When two neighboring edge segments take the identical labels, we have no cost. Otherwise, the cost is a positive value $\beta.\mu(i,j)\geq 0$. Function $\mu(i,j)$ is computed in equation \eqref{eq17}.
	
    \begin{equation}
     \mu(i,j)=A_{\text{exc}}(i,j).e^{-\frac{d_{i,j}}{\sigma_1}}.e^{-\frac{|f_j-f_i|}{\sigma_2}} \label{eq17}
    \end{equation}
    According to equation \eqref{eq17}, more excitation causes more cost for different labeling. In other words, when edge segments $i$ and $j$ excite each other, they tend to be on the same contour and they probably have similar labels. Therefore, both belong to either a strong or weak contour and both  should be either maintained or removed. According to equation \eqref{eq17}, the amount of excitation is modulated by the difference in strength.
    
    \subsection{Inference and Optimization }
    \label{chp.opt}
     
     The energy and probability of each labeling assignment can be computed by equations \eqref{eq8} and \eqref{eq7}, respectively. Our final goal is to find the labeling $x^{*}$ that has the maximum probability:
      	    \begin{align}	  
	    \begin{split}
	    x^{*} &=\arg\max_{x}\  p(x_{1}, x_{2}, ..., x_{n} | C(I)) \\
	          &=\arg\max_{x} \ln\bigl( p(x_{1}, x_{2}, ..., x_{n} | C(I))\bigr)\\
	          &=\arg\max_{x} -E(x_{1}, x_{2}, ..., x_{n}|C(I))-Z(C(I))
	    \end{split}  \label{eq18}
	    \end{align}
	  Given image I, partition function $Z(C(I))$ is constant and we should only minimize energy function defined in equation \eqref{eq8}. Fortunately, suggested pairwise energy specified in equation \eqref{eq14} satisfies sub-modularity property:
	  
      	    \begin{equation}
	    \epsilon_{i,j}{(0,0)}+\epsilon_{i,j}{(1,1)} \leq \epsilon_{i,j}{(0,1)} + \epsilon_{i,j}{(1,0)} \\, \forall i,j \label{eq6} 
	    \end{equation} 
	    In these circumstances, the MAP inference can be reduced to a min-cut problem  \cite{jegelka2011submodularity,kolmogorov2004energy} and the optimal $x^{*}$ is achieved very fast. In other words, we can design a graph so that the source and sink nodes demonstrate labels 0 and 1, respectively. Each s-t cut corresponds to a specific assignment and its cost to the energy of the assignment given in equation \eqref{eq8}. Consequently, the optimal assignment corresponds to a minimum cut
	    
      Since the performance of this framework strongly depends on the formation of energy function of equation \eqref{eq8}, parameter estimation has a very important role to play in this regard. Suppose the set $\theta$ consists of all parameters of energy functions and we have a training set  $T=\{C(I^i),B^i{\}}_{i=1}^N$. The set  $C(I^i)$ is edge segments of image  $I^i$ and the set $B^i$ is their corresponding binary labels. The common way here is to find the maximum likelihood estimation. Likelihood function is defined in equation \eqref{likelihood}. In order to maximize L, we should compute partition function Z for different $\theta$s which is necessary but practically impossible.
      \begin{equation}   
L(\theta;T)=\prod_{i=1}^{N}\frac{1}{Z(C(I^i),\theta)}e^{-E(B^i|C(I^i);\theta)} \label{likelihood}
      \end{equation}

    Many and various solutions have been offered to estimate the likelihood function. As an illustration, we can compute pseudo-likelihood, instead. Yet, this estimation reduces the performance of the framework. In other words, raising the pseudo-likelihood  does not necessarily reduce the contour integration error. On the other hand, in addition to energy parameters $\theta$, there exist some effective structural parameters $\theta_{s}$ like
    Markov blanket size. It is crystal clear that we cannot put pseudo-likelihood to use so as to optimize these parameters.
    
    
    To overcome the aforementioned problems and enhance the performance, we decided not to put gradient-based methods to use and optimize the output of the framework directly. The nature of the inference in our framework being swift makes this solution feasible. After defining an evaluation function which gets the parameter values and runs the framework on all training sets, an average F-measure is computed. Then, we make use of the fast and powerful MCS \cite{huyer1999global} global optimization to maximize this function.
    

    \section{Results}
		\label{chp.result}

    In this section, we favore the  proposed approach to improve other methods. Our framework gets an arbitrary soft-edge map, and creates an improved binary contour map through employing contextual influences. Applying our method on a binary synthetic edge map, we observe that the smooth contour  popps out of the background noise.
    
    \subsection{Graph-cut vs Thresholding}
    
    Many contour and edge detection algorithms provide us with a soft-value as an output. The higher values mean the higher likelihood of being a part of the contour. Normally, a threshold is set to convert the soft-value image to a contour image. In this subsection, in order for us to evaluate the performance improvement achieved, we prefer our method over simple thresholding of the resulting soft-values.

    
    Setting an overall threshold to achieve the most flawless performance for certain objectives (F-measure here) has a number of drawbacks. First, the optimal threshold for each image can be different by comparison with the best overall threshold. Furthermore, other factors should also be considered so as to transform the soft values into the binary image. Thus the following factors of smoothness, length of contour and continuity should be taken into consideration, for instance. The framework we have proposed, can be utilized for transforming soft value outputs of a given method into binary images, incorporating factors used for contour integration by humans. Distinct steps of the algorithm for to utilize our framework for contour integration are as follows:
    
    
   \begin{enumerate}
    	
    	\item Detecting initial edge segments by putting a threshold $\text{th}_0$ on the soft values. The pixels that are higher than this threshold are considered as edge segments. The descriptor for each segment is the soft value for that pixel.
 
    	
    	\item A binary image is computed by thresholding using the value of the previous step. The orientation of the edge segments are computed from the binary image employing the skeleton orientation method\footnote{http://github.com/tsogkas/matlab-utils/blob/master/skeletonOrientation.m}.

    	
    	\item Using the edge segments descriptor, position and orientation, we construct a CRF for contour integration making use of our proposed framework in Section \ref{chp.method}. 

    	
    	\item Using the maximum a posteriori inference for computing the best contour integration. This is solved by the graph cut in our framework (see Subsection  \ref{chp.opt}).

    \end{enumerate}

        \begin{figure*} [thb]	
    	\begin{subfigure}[b]{0.16\textwidth}
    		\includegraphics[width=\textwidth]{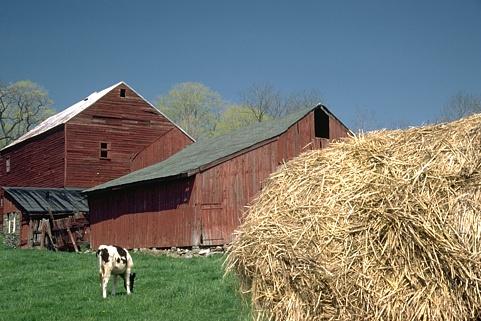}
    	\end{subfigure}
    	\begin{subfigure}[b]{0.16\textwidth}
    		\includegraphics[width=\textwidth]{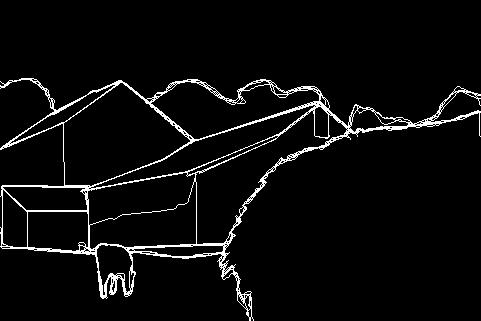}
    	\end{subfigure}
    	\begin{subfigure}[b]{0.16\textwidth}
    		\includegraphics[width=\textwidth]{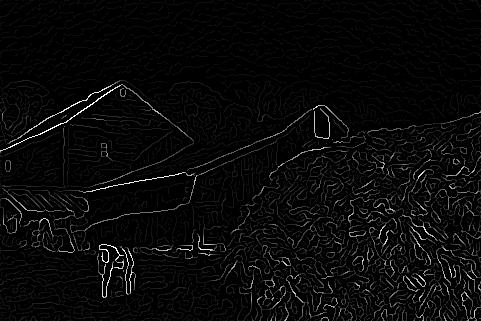}
    	\end{subfigure}
    	\begin{subfigure}[b]{0.16\textwidth}
    		\includegraphics[width=\textwidth]{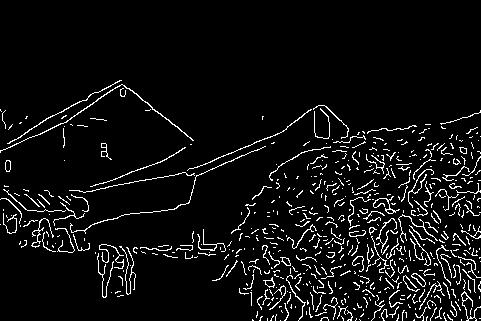}
    	\end{subfigure}
    	\begin{subfigure}[b]{0.16\textwidth}
    		\includegraphics[width=\textwidth]{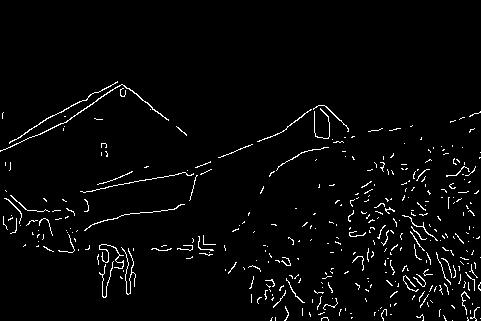}
    	\end{subfigure}
    	\begin{subfigure}[b]{0.16\textwidth}
    		\includegraphics[width=\textwidth]{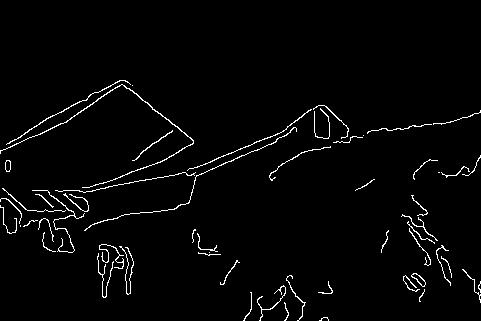}
    	\end{subfigure}
    	\begin{subfigure}[b]{0.16\textwidth}
    		\includegraphics[width=\textwidth]{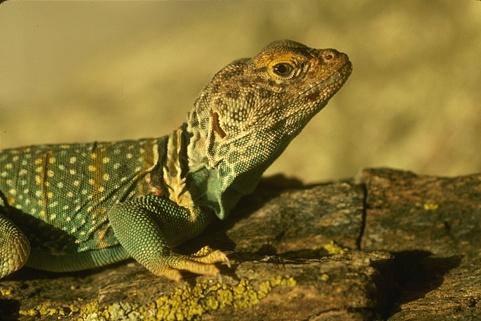}
    	\end{subfigure}
    	\begin{subfigure}[b]{0.16\textwidth}
    		\includegraphics[width=\textwidth]{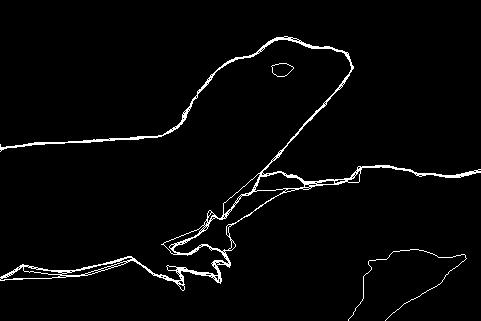}
    	\end{subfigure}
    	\begin{subfigure}[b]{0.16\textwidth}
    		\includegraphics[width=\textwidth]{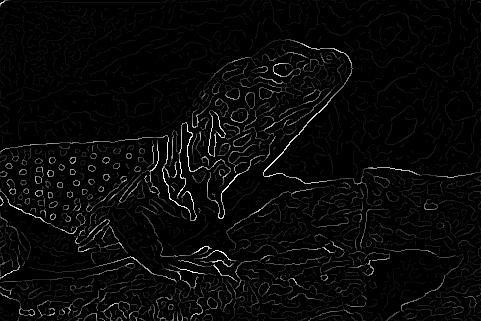}
    	\end{subfigure}
    	\begin{subfigure}[b]{0.16\textwidth}
    		\includegraphics[width=\textwidth]{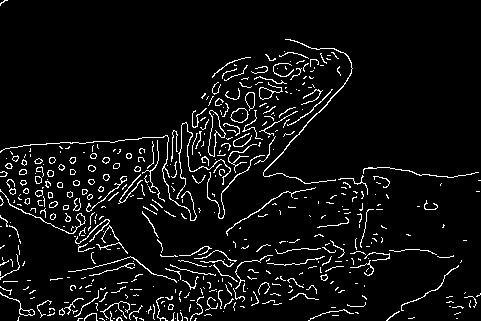}
    	\end{subfigure}
    	\begin{subfigure}[b]{0.16\textwidth}
    		\includegraphics[width=\textwidth]{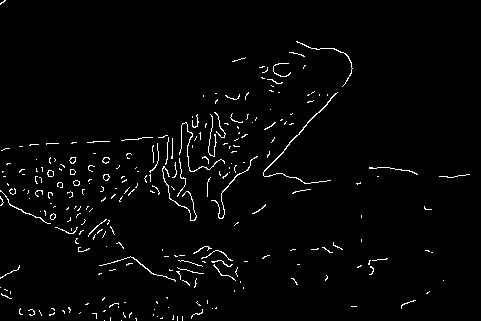}
    	\end{subfigure}
    	\begin{subfigure}[b]{0.16\textwidth}
    		\includegraphics[width=\textwidth]{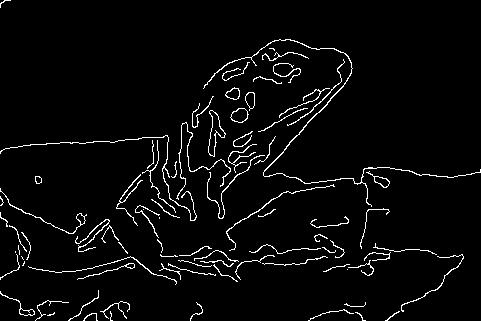}
    	\end{subfigure}
    	\begin{subfigure}[b]{0.16\textwidth}
    		\includegraphics[width=\textwidth]{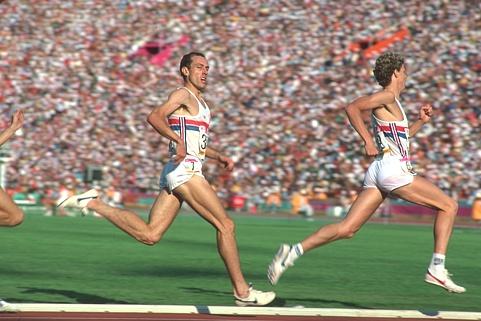}
    	\end{subfigure}
    	\begin{subfigure}[b]{0.16\textwidth}
    		\includegraphics[width=\textwidth]{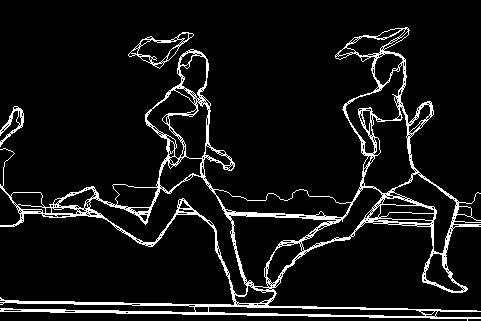}
    	\end{subfigure}
    	\begin{subfigure}[b]{0.16\textwidth}
    		\includegraphics[width=\textwidth]{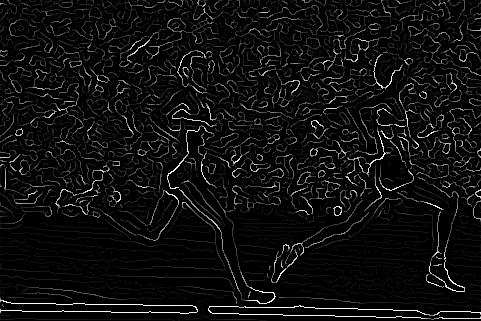}
    	\end{subfigure}
    	\begin{subfigure}[b]{0.16\textwidth}
    		\includegraphics[width=\textwidth]{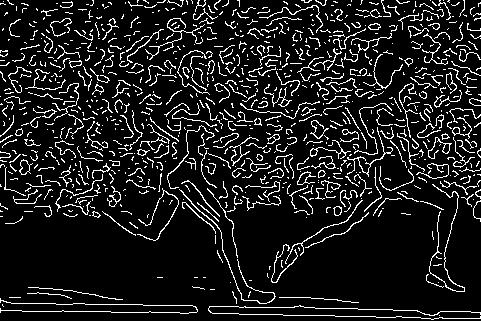}
    	\end{subfigure}
    	\begin{subfigure}[b]{0.16\textwidth}
    		\includegraphics[width=\textwidth]{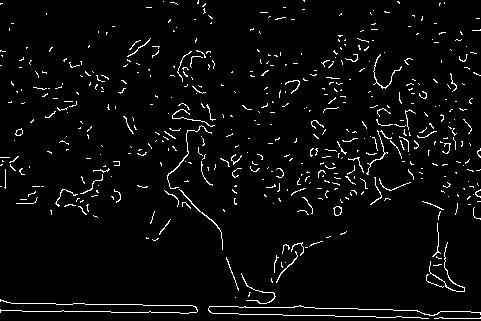}
    	\end{subfigure}
    	\begin{subfigure}[b]{0.16\textwidth}
    		\includegraphics[width=\textwidth]{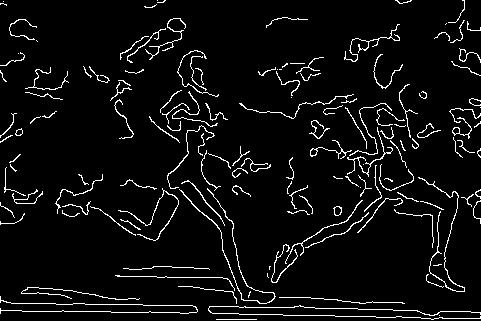}
    	\end{subfigure}
    	\begin{subfigure}[b]{0.16\textwidth}
    		\includegraphics[width=\textwidth]{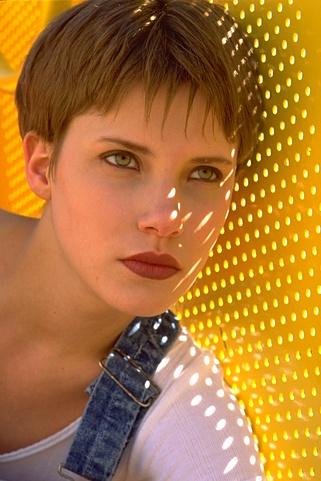}
    		\caption{}
    	\end{subfigure}
    	\begin{subfigure}[b]{0.16\textwidth}
    		\includegraphics[width=\textwidth]{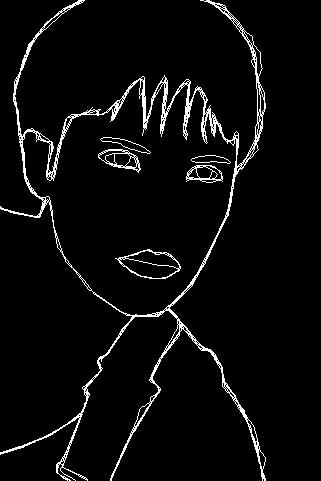}
    		\caption{}
    	\end{subfigure}
    	\begin{subfigure}[b]{0.16\textwidth}
    		\includegraphics[width=\textwidth]{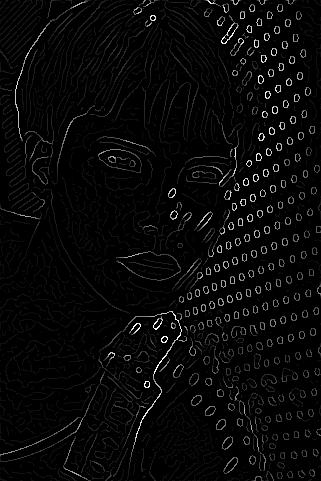}
    		\caption{}
    	\end{subfigure}
    	\begin{subfigure}[b]{0.16\textwidth}
    		\includegraphics[width=\textwidth]{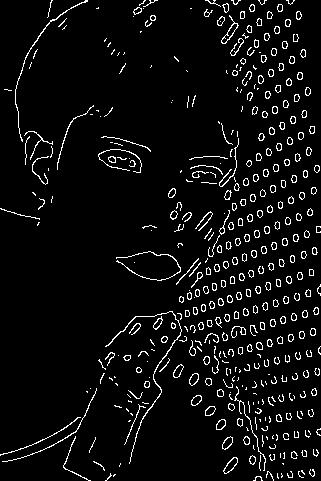}
    		\caption{}
    	\end{subfigure}
    	\begin{subfigure}[b]{0.16\textwidth}
    		\includegraphics[width=\textwidth]{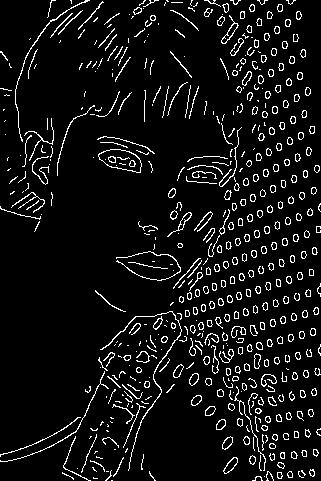}
    		\caption{}
    	\end{subfigure}
    	\begin{subfigure}[b]{0.16\textwidth}
    		\includegraphics[width=\textwidth]{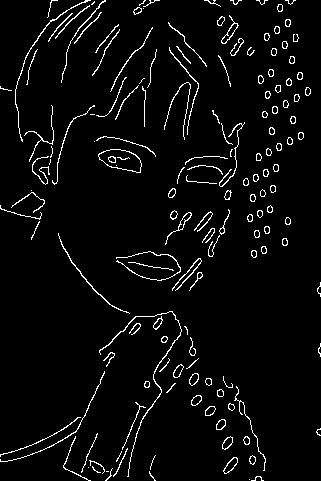}
    		\caption{}
    	\end{subfigure}
    	\caption{Evaluation of the proposed method on the gradient magnitude algorithm. (a) Original image. (b) Ground truth image. (c) Gradient magnitude soft output. The binary outputs are computed by three manners:  (d) Applying the average optimal threshold (e) Applying the optimal threshold for image (f) Applying the proposed method. \label{fig:GM}}    
    \end{figure*}

                \begin{figure*} [t!]	
    	\begin{subfigure}[b]{0.16\textwidth}
    		\includegraphics[width=\textwidth]{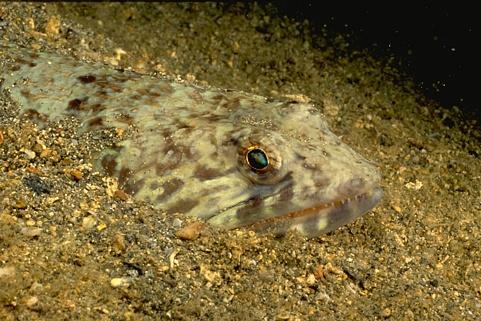}
    	\end{subfigure}
    	\begin{subfigure}[b]{0.16\textwidth}
    		\includegraphics[width=\textwidth]{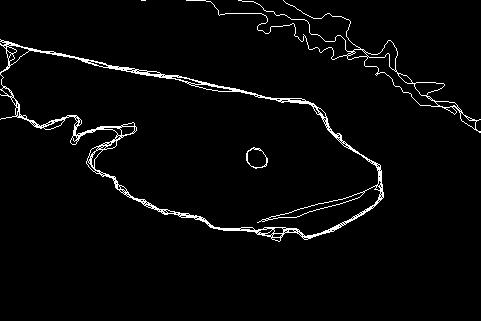}
    	\end{subfigure}
    	\begin{subfigure}[b]{0.16\textwidth}
    		\includegraphics[width=\textwidth]{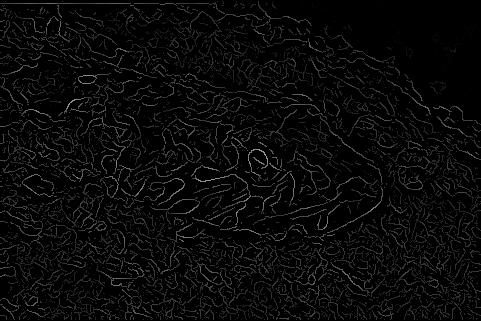}
    	\end{subfigure}
    	\begin{subfigure}[b]{0.16\textwidth}
    		\includegraphics[width=\textwidth]{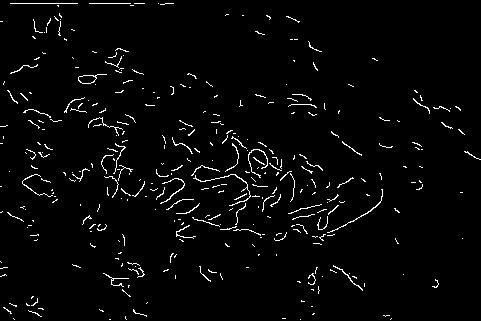}
    	\end{subfigure}
    	\begin{subfigure}[b]{0.16\textwidth}
    		\includegraphics[width=\textwidth]{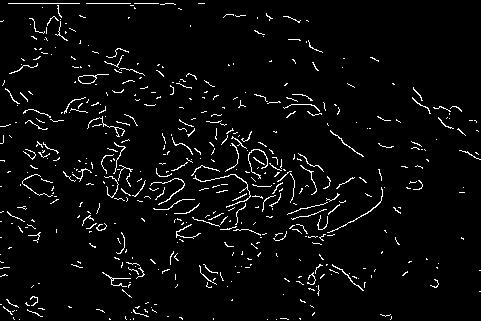}
    	\end{subfigure}
    	\begin{subfigure}[b]{0.16\textwidth}
    		\includegraphics[width=\textwidth]{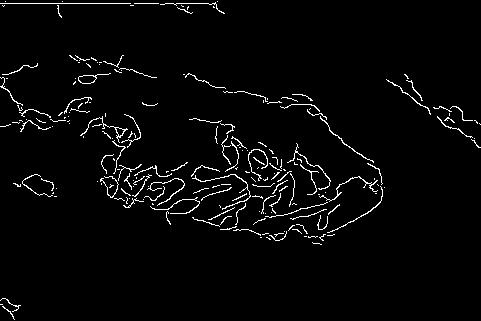}
    	\end{subfigure}
    	\begin{subfigure}[b]{0.16\textwidth}
    		\includegraphics[width=\textwidth]{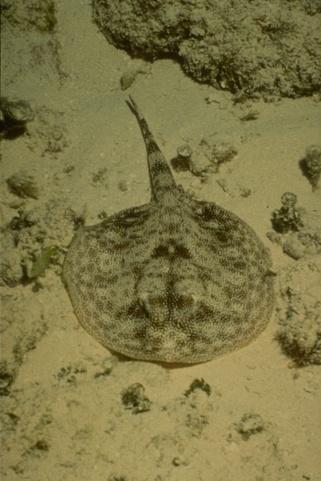}
    	\end{subfigure}
    	\begin{subfigure}[b]{0.16\textwidth}
    		\includegraphics[width=\textwidth]{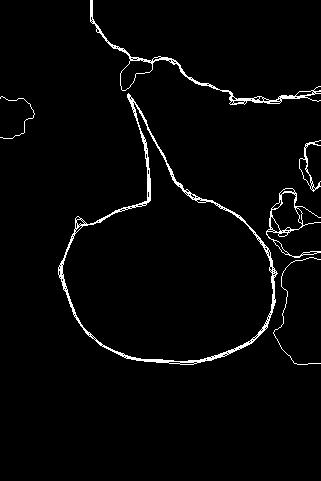}
    	\end{subfigure}
    	\begin{subfigure}[b]{0.16\textwidth}
    		\includegraphics[width=\textwidth]{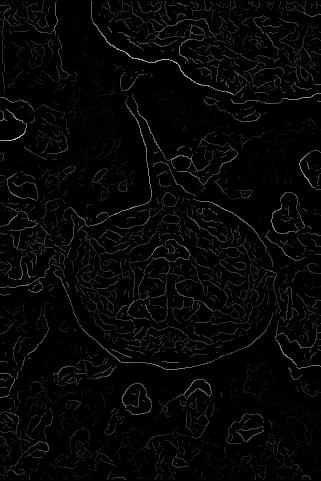}
    	\end{subfigure}
    	\begin{subfigure}[b]{0.16\textwidth}
    		\includegraphics[width=\textwidth]{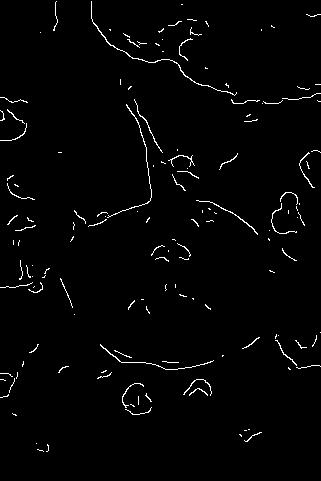}
    	\end{subfigure}
    	\begin{subfigure}[b]{0.16\textwidth}
    		\includegraphics[width=\textwidth]{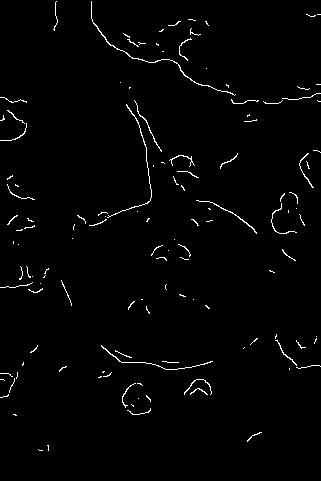}
    	\end{subfigure}
    	\begin{subfigure}[b]{0.16\textwidth}
    		\includegraphics[width=\textwidth]{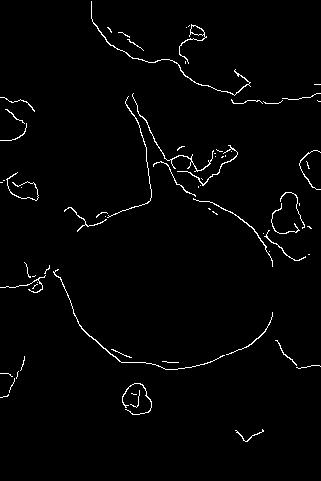}
    	\end{subfigure}
    	\begin{subfigure}[b]{0.16\textwidth}
    		\includegraphics[width=\textwidth]{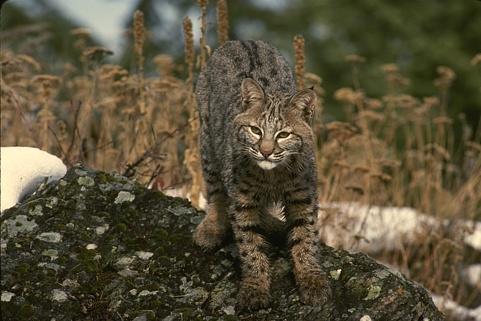}
    		\caption{}
    	\end{subfigure}
    	\begin{subfigure}[b]{0.16\textwidth}
    		\includegraphics[width=\textwidth]{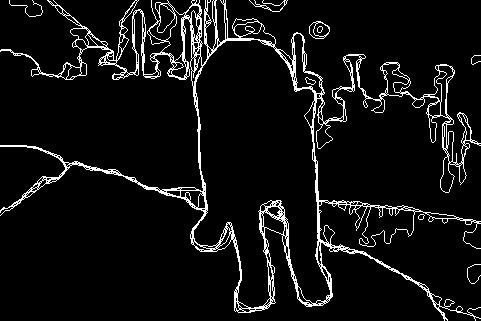}
    		\caption{}
    	\end{subfigure}
    	\begin{subfigure}[b]{0.16\textwidth}
    		\includegraphics[width=\textwidth]{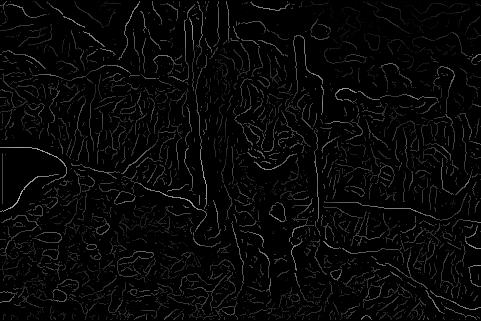}
    		\caption{}
    	\end{subfigure}
    	\begin{subfigure}[b]{0.16\textwidth}
    		\includegraphics[width=\textwidth]{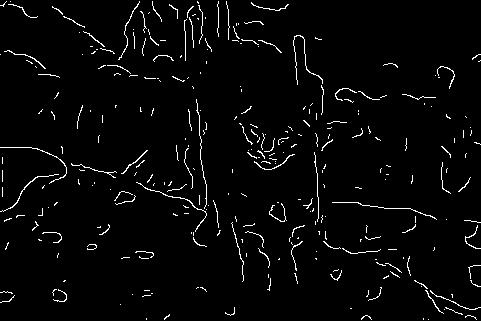}
    		\caption{}
    	\end{subfigure}
    	\begin{subfigure}[b]{0.16\textwidth}
    		\includegraphics[width=\textwidth]{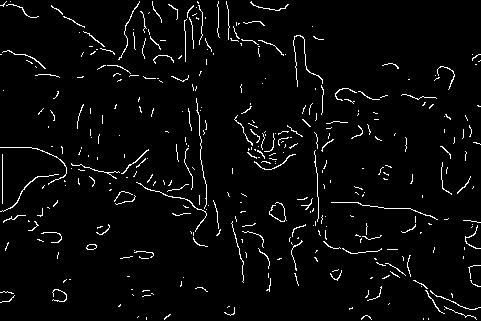}
    		\caption{}
    	\end{subfigure}
    	\begin{subfigure}[b]{0.16\textwidth}
    		\includegraphics[width=\textwidth]{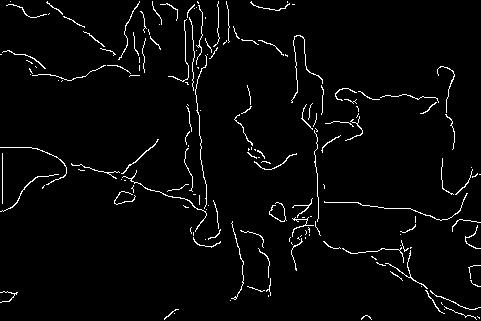}
    		\caption{}
    	\end{subfigure}
    	\caption{Evaluation of the proposed method on the mPb algorithm  \cite{arbelaez2011contour}. Columns description is the same as Figure \ref{fig:GM}. \label{fig:mpb}}    
    \end{figure*} 
   
    \begin{figure*} [t!]	
    	\begin{subfigure}[b]{0.16\textwidth}
    		\includegraphics[width=\textwidth]{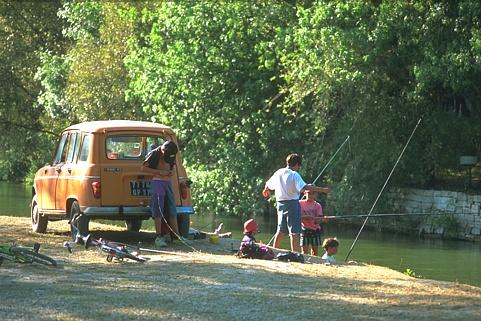}
    	\end{subfigure}
    	\begin{subfigure}[b]{0.16\textwidth}
    		\includegraphics[width=\textwidth]{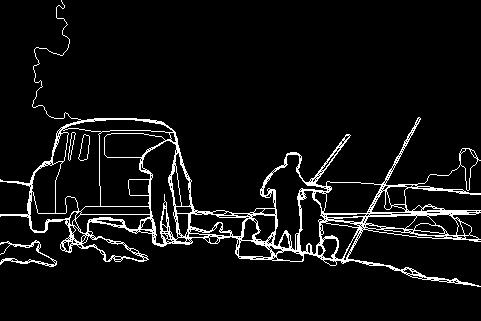}
    	\end{subfigure}
    	\begin{subfigure}[b]{0.16\textwidth}
    		\includegraphics[width=\textwidth]{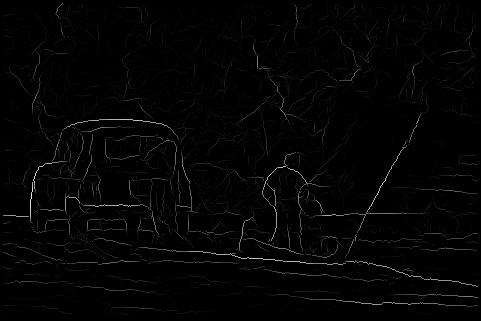}
    	\end{subfigure}
    	\begin{subfigure}[b]{0.16\textwidth}
    		\includegraphics[width=\textwidth]{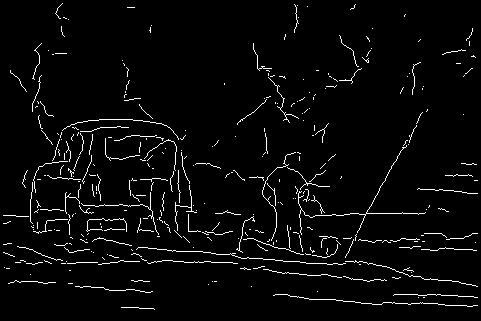}
    	\end{subfigure}
    	\begin{subfigure}[b]{0.16\textwidth}
    		\includegraphics[width=\textwidth]{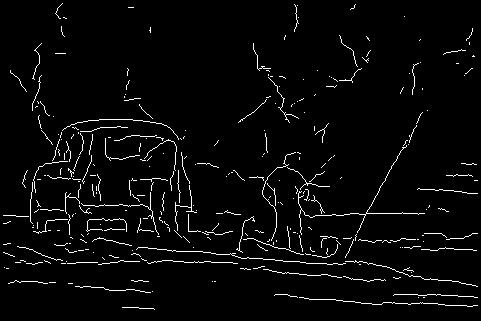}
    	\end{subfigure}
    	\begin{subfigure}[b]{0.16\textwidth}
    		\includegraphics[width=\textwidth]{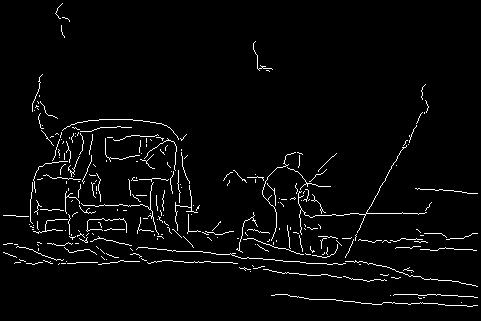}
    	\end{subfigure}
    	\begin{subfigure}[b]{0.16\textwidth}
    		\includegraphics[width=\textwidth]{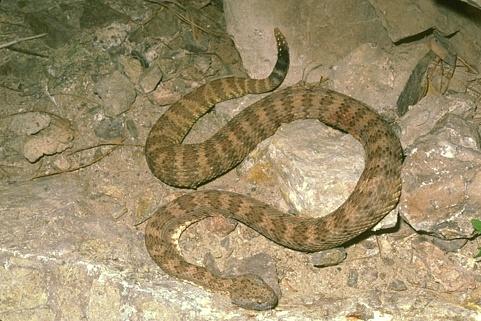}
    	\end{subfigure}
    	\begin{subfigure}[b]{0.16\textwidth}
    		\includegraphics[width=\textwidth]{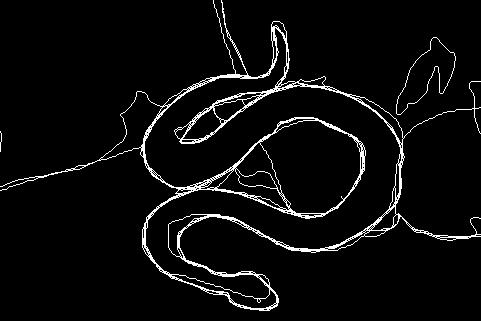}
    	\end{subfigure}
    	\begin{subfigure}[b]{0.16\textwidth}
    		\includegraphics[width=\textwidth]{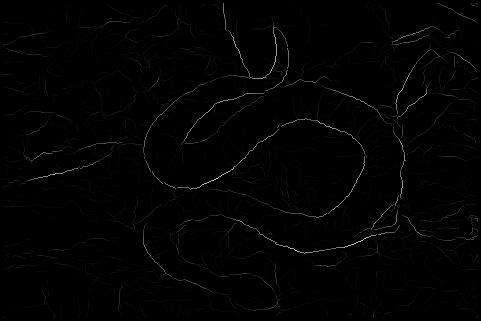}
    	\end{subfigure}
    	\begin{subfigure}[b]{0.16\textwidth}
    		\includegraphics[width=\textwidth]{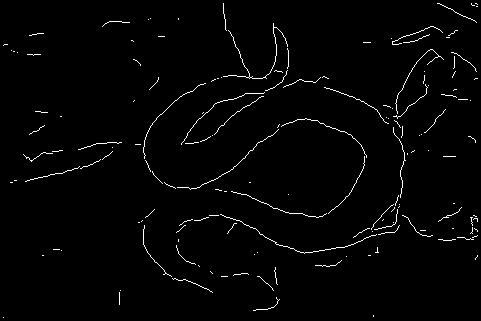}
    	\end{subfigure}
    	\begin{subfigure}[b]{0.16\textwidth}
    		\includegraphics[width=\textwidth]{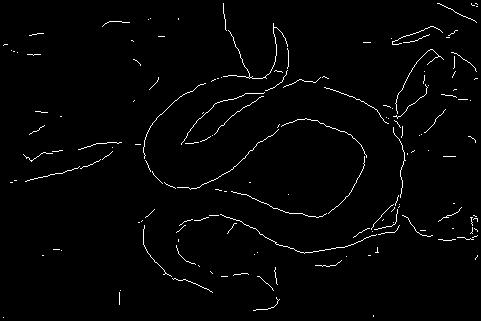}
    	\end{subfigure}
    	\begin{subfigure}[b]{0.16\textwidth}
    		\includegraphics[width=\textwidth]{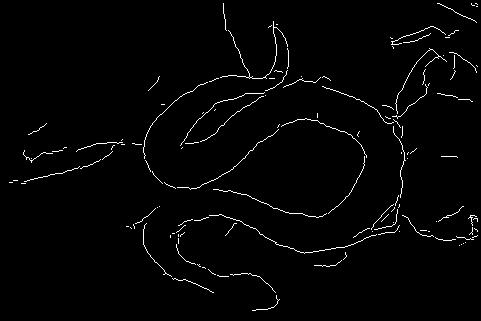}
    	\end{subfigure}
    	\begin{subfigure}[b]{0.16\textwidth}
    		\includegraphics[width=\textwidth]{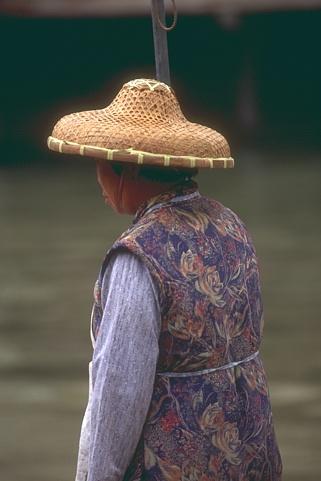}
    		\caption{}
    	\end{subfigure}
    	\begin{subfigure}[b]{0.16\textwidth}
    		\includegraphics[width=\textwidth]{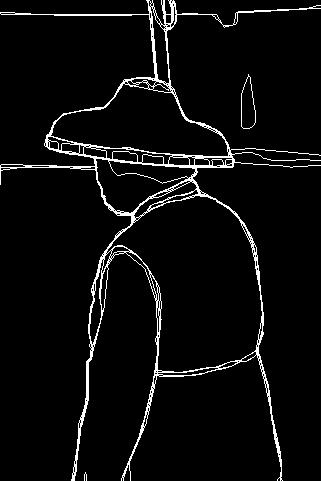}
    		\caption{}
    	\end{subfigure}
    	\begin{subfigure}[b]{0.16\textwidth}
    		\includegraphics[width=\textwidth]{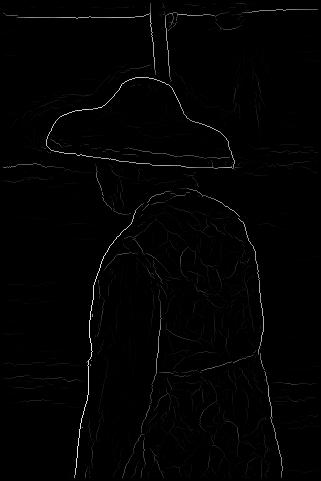}
    		\caption{}
    	\end{subfigure}
    	\begin{subfigure}[b]{0.16\textwidth}
    		\includegraphics[width=\textwidth]{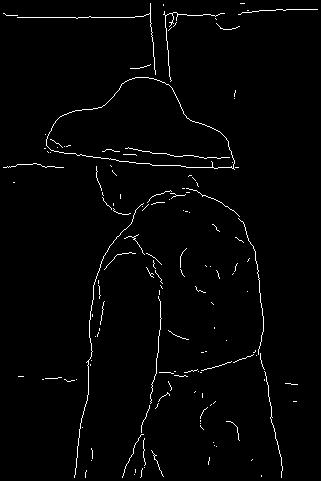}
    		\caption{}
    	\end{subfigure}
    	\begin{subfigure}[b]{0.16\textwidth}
    		\includegraphics[width=\textwidth]{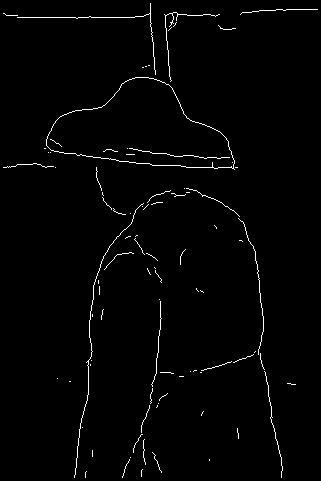}
    		\caption{}
    	\end{subfigure}
    	\begin{subfigure}[b]{0.16\textwidth}
    		\includegraphics[width=\textwidth]{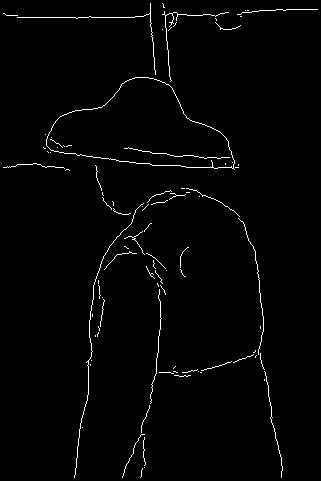}
    		\caption{}
    	\end{subfigure}
    	\caption{Evaluation of the proposed method on the SCG algorithm  \cite{xiaofeng2012discriminatively}. Columns description is the same as Figure \ref{fig:GM}.\label{fig:x}}    
    \end{figure*}

    We consider Markov blanket $N_{i}$ as $n\times n$ square centered at the position of the edge segment $i$. For each contour detection method to be improved, structure parameters ${\{n,t_0\}}$ as well as energy parameters should be tuned. In this regard, the BSDS500 dataset  \cite{arbelaez2011contour} is used  for the parameters to be trained. As we mentioned in Subsection  \ref{chp.opt}, the average F-measure is maximized by using global MCS  \cite{huyer1999global} optimization.

    In order that the proposed method is evaluated qualitatively, the soft output of the three methods is taken into account: gradient magnitude, multi scale PB  \cite{arbelaez2011contour} and SCG  \cite{xiaofeng2012discriminatively}. They are respectively low, mid and high level descriptors. You can see the evaluation results in figures \ref{fig:GM},\ref{fig:mpb} and \ref{fig:x}. In these figures, columns b and c depict the ground truth and soft map, respectively. We utilized two thresholds to convert the soft map to the binary output. The first threshold is the average optimal one on dataset, and the second is the optimal threshold for the image. These outputs are represented in columns d and e. Column f indicates the result of the proposed method. As you can see, the output of our framework is more meaningful than that of the two others. In comparison to thresholding, our method is capable of removing more texture noise and creating more continuous contours.
    

         \begin{table}[t]
         \begin{center}
    	\begin{tabular}{ |c||c|c|  }
    		\hline
    		& thresholding  & proposed method \\
    		\hline
    		GM & 0.6 & 0.63\\
    		mPb  \cite{arbelaez2011contour} & 0.69 & 0.7\\
    		gPb  \cite{arbelaez2011contour}& 0.71 & 0.72\\
    		SCG  \cite{xiaofeng2012discriminatively} & 0.74 & 0.75\\
    		\hline
    	\end{tabular}
	\end{center}
    	\caption{Quantitative evaluation of the proposed method on the BSDS500 dataset  \cite{arbelaez2011contour}. The average ODS F-measure is computed.\label{tab:BSDS500}}
    \end{table}
    
    Applying surround modulation assists us to highlight the main contour and remove the texture noise better. Our method has been especially more effective on low and mid-level descriptors in comparison to the higher ones.  In table~\ref{tab:BSDS500}, ODS F-measure  \cite{arbelaez2011contour} is computed based on the BSDS500 dataset. Taking a look at the quantitative evaluation in this table, it is clear that by replacing thresholding with our graph-based framework, the performance of not only low-level and mid-level soft-values but also high-level ones have been improved.

        \begin{figure}[t]
 	
 	\begin{subfigure}[b]{0.49\textwidth}
 		\includegraphics[width=\textwidth]{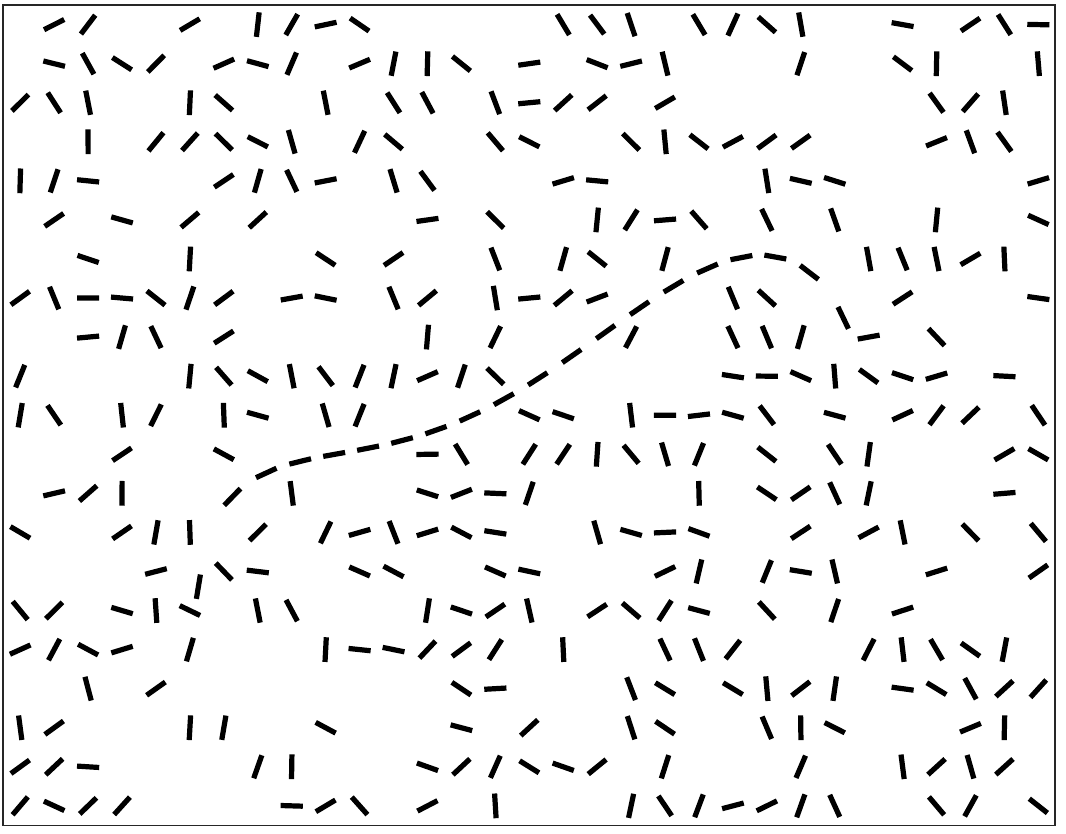}
 		\caption{ \label{fig:psychoa}}
 	\end{subfigure}
 	\begin{subfigure}[b]{0.49\textwidth}
 		\includegraphics[width=\textwidth]{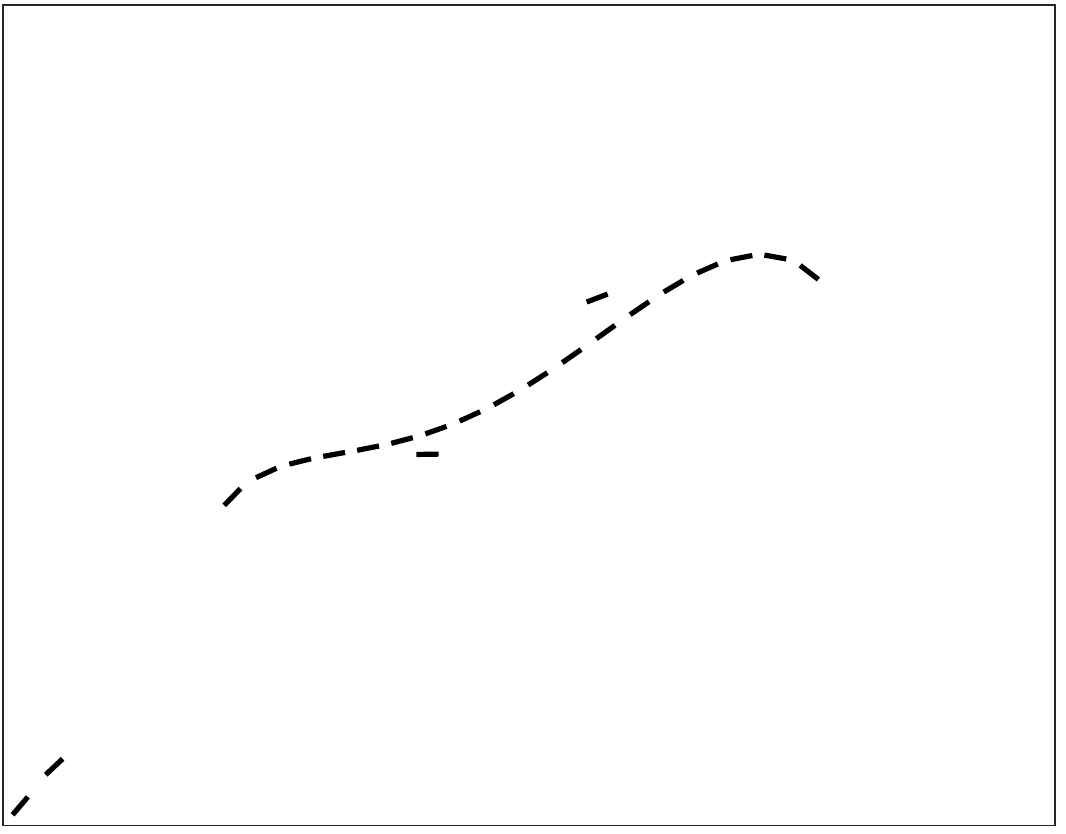}
 		\caption{ \label{fig:psychob}}
 	\end{subfigure}
 	\caption{Evaluation of the proposed method on a synthetic binary image. (a) A synthetic binary image was created. (b) Similar to human contour integration, our framework could pop out the smooth contour located among the random diffused edge segments.  \label{fig:psycho}}
 \end{figure}
      \subsection{Visual grouping}
      This subsection is aimed at testing the proposed framework of ours using a psychophysical experiment. In this respect, a smooth contour was located among random diffused edge segments as presented in figure \ref{fig:psychoa}. The output of the proposed method is illustrated in figure \ref{fig:psychob}. As you can see, most of the noise in the background has been removed and the only thing to remain is the smooth contour. In this case, the edge segments have similar descriptors, and both smoothness and length are among the most influential factors. The edge segments located on smooth contour excite each other, and lead to a rise in the chance of selection. Meanwhile, the background noise and isolated edge segments are suppressed, as well.
      
      
      Similarly when a human look at this image, can easily pop out embedded smooth contour. Now you can easily understand the importance of the second logistic function in unary energy of equation \eqref{eq10}.  unless this term is not taken into account, all contour segments are on a par with each other regarding the unary energy level. According to pairwise energy in equation \eqref{eq14}, getting equal labels has no cost. Therefore, energy function of equation~\eqref{eq8} is minimized when the entire contour segments belong to the same class.
      
      \section{Discussion and Conclusions}

  There are multitude of different approaches to contour detection and edge detection, and most of them give a soft-value as an output before applying a final threshold. In this paper, we proposed a graph-based framework that gets the soft-value of other methods as its input and creates more meaningful contours. Inspired by the concept of non-classical receptive fields in the primary visual cortex, we considered important factors such as connectivity, smoothness, and length of the contour beside the soft-values. Edge segments with higher soft values have more chances to be selected. On the other hand, edge segments which form smooth contours excite each other and increase selection chance. As well as, background noises and isolated edge segments are suppressed.

  In our graph-based framework, contours were detected by solving a maximum a posteriori problem. This resulted in labeling of edge segments that gives the highest probability or lowest energy. We designed the framework such that this optimization problem is reduced to graph-cut, and so the optimal solution can be achieved very fast without any iterative steps. Fast inference helped us to use MCS  \cite{huyer1999global} global optimization to train parameters by maximizing average F-measure on training data. This means we directly maximized the accuracy of contour integration instead of pseudo-likelihood function.

  Our method had better performance than applying an optimal threshold especially on low-level and mid-level soft values. For example, the performance of gradient magnitude was improved from 0.6 to 0.63. In the last experiment, we showed that even if all edge segments have the same values, our method can pop out smooth contours from background noise. This is the strong point of our approach which can remove background noise in binary edge-map by applying surround modulation, while other popular methods such as Canny cannot.
 \bibliography{scholar} 

\begin{thebibliography}{10}

\bibitem{arbelaez2011contour}
P.~Arbelaez, M.~Maire, C.~Fowlkes, and J.~Malik, ``Contour detection and
  hierarchical image segmentation,'' {\em IEEE Transactions on Pattern Analysis
  and Machine Intelligence}, vol.~33, no.~5, pp.~898--916, 2010.
\newblock doi: https://doi.org/10.1109/TPAMI.2010.161.

\bibitem{xiaofeng2012discriminatively}
R.~Xiaofeng and L.~Bo, ``Discriminatively trained sparse code gradients for
  contour detection,'' in {\em Advances in Neural Information Processing
  Systems 25} (F.~Pereira, C.~J.~C. Burges, L.~Bottou, and K.~Q. Weinberger,
  eds.), pp.~584--592, Curran Associates, Inc., 2012.

\bibitem{canny1986computational}
J.~Canny, ``A computational approach to edge detection,'' {\em IEEE
  Transactions on Pattern Analysis and Machine Intelligence}, vol.~PAMI-8,
  no.~6, pp.~679--698, 1986.
\newblock doi: https://doi.org/10.1109/TPAMI.1986.4767851.

\bibitem{grigorescu2003contour}
C.~Grigorescu, N.~Petkov, and M.~A. Westenberg, ``Contour detection based on
  nonclassical receptive field inhibition,'' {\em IEEE Transactions on Image
  Processing}, vol.~12, no.~7, pp.~729--739, 2003.
\newblock doi: https://doi.org/10.1109/TIP.2003.814250.

\bibitem{petkov2003suppression}
N.~Petkov and M.~A. Westenberg, ``Suppression of contour perception by
  band-limited noise and its relation to nonclassical receptive field
  inhibition,'' {\em Biological Cybernetics}, vol.~88, no.~3, pp.~236--246,
  2003.
\newblock doi: https://doi.org/10.1007/s00422-002-0378-2.

\bibitem{grigorescu2004contour}
C.~Grigorescu, N.~Petkov, and M.~A. Westenberg, ``Contour and boundary
  detection improved by surround suppression of texture edges,'' {\em Image and
  Vision Computing}, vol.~22, no.~8, pp.~609--622, 2004.
\newblock doi: https://doi.org/10.1016/j.imavis.2003.12.004.

\bibitem{papari2007biologically}
G.~Papari, P.~Campisi, N.~Petkov, and A.~Neri, ``A biologically motivated
  multiresolution approach to contour detection,'' {\em EURASIP Journal on
  Advances in Signal Processing}, 2007.
\newblock doi: https://doi.org/10.1155/2007/71828.

\bibitem{zeng2011center}
C.~Zeng, Y.~Li, and C.~Li, ``Center--surround interaction with adaptive
  inhibition: A computational model for contour detection,'' {\em NeuroImage},
  vol.~55, no.~1, pp.~49--66, 2011.
\newblock doi: https://doi.org/10.1016/j.neuroimage.2010.11.067.

\bibitem{wei2013contour}
H.~Wei, B.~Lang, and Q.~Zuo, ``Contour detection model with multi-scale
  integration based on non-classical receptive field,'' {\em Neurocomputing},
  vol.~103, pp.~247--262, 2013.
\newblock doi: https://doi.org/10.1016/j.neucom.2012.09.027.

\bibitem{zeng2011contour}
C.~Zeng, Y.~Li, K.~Yang, and C.~Li, ``Contour detection based on a
  non-classical receptive field model with butterfly-shaped inhibition
  subregions,'' {\em Neurocomputing}, vol.~74, no.~10, pp.~1527--1534, 2011.
\newblock doi: https://doi.org/10.1016/j.neucom.2010.12.022.

\bibitem{li1998neural}
Z.~Li, ``A neural model of contour integration in the primary visual cortex,''
  {\em Neural Computation}, vol.~10, no.~4, pp.~903--940, 1998.
\newblock doi: https://doi.org/10.1162/089976698300017557.

\bibitem{sang2017contour}
Q.~Sang, B.~Cai, and H.~Chen, ``Contour detection improved by context-adaptive
  surround suppression,'' {\em PloS ONE}, vol.~12, no.~7, 2017.
\newblock doi: https://doi.org/10.1371/journal.pone.0181792.

\bibitem{su2017contour}
P.~Su, X.~Ren, and J.~Ma, ``Contour detection model based on the combination of
  surround facilitation and inhibition,'' in {\em 2nd International Conference
  on Multimedia and Image Processing}, pp.~32--37, 2017.
\newblock doi: https://doi.org/10.1109/ICMIP.2017.14.

\bibitem{guy1992perceptual}
G.~Guy and G.~Medioni, ``Perceptual grouping using global saliency-enhancing
  operators,'' in {\em 11th International Conference on Pattern Recognition},
  pp.~99--103, 1992.
\newblock doi: https://doi.org/10.1109/ICPR.1992.201517.

\bibitem{hess2014contour}
R.~F. Hess, K.~A. May, and S.~O. Dumoulin, ``Contour integration:
  Psychophysical, neurophysiological, and computational perspectives,'' in {\em
  Oxford Handbook of Perceptual Organization} (J.~Wagemans, ed.), pp.~189--206,
  Oxford, U.K.: Oxford University Press, 2015.
\newblock doi: https://doi.org/10.1093/oxfordhb/9780199686858.013.013.

\bibitem{kuai2017contour}
S.-G. Kuai, W.~Li, C.~Yu, and Z.~Kourtzi, ``Contour integration over time:
  Psychophysical and fmri evidence,'' {\em Cerebral Cortex}, vol.~27, no.~5,
  pp.~3042--3051, 2017.
\newblock doi: https://doi.org/10.1093/cercor/bhw147.

\bibitem{qiu2016responses}
C.~Qiu, P.~C. Burton, D.~Kersten, and C.~A. Olman, ``Responses in early visual
  areas to contour integration are context dependent,'' {\em Journal of
  Vision}, vol.~16, no.~8, pp.~1--18, 2016.
\newblock doi: https://doi.org/10.1167/16.8.19.

\bibitem{hubel1962receptive}
D.~H. Hubel and T.~N. Wiesel, ``Receptive fields, binocular interaction and
  functional architecture in the cat's visual cortex,'' {\em The Journal of
  Physiology}, vol.~160, no.~1, pp.~106--154, 1962.
\newblock doi: https://doi.org/10.1113/jphysiol.1962.sp006837.

\bibitem{hubel1968receptive}
D.~H. Hubel and T.~N. Wiesel, ``Receptive fields and functional architecture of
  monkey striate cortex,'' {\em The Journal of Physiology}, vol.~195, no.~1,
  pp.~215--243, 1968.
\newblock doi: https://doi.org/10.1113/jphysiol.1968.sp008455.

\bibitem{hess2003contour}
R.~Hess, A.~Hayes, and D.~Field, ``Contour integration and cortical
  processing,'' {\em Journal of Physiology-Paris}, vol.~97, no.~2-3,
  pp.~105--119, 2003.
\newblock doi: https://doi.org/10.1016/j.jphysparis.2003.09.013.

\bibitem{kapadia2000spatial}
M.~K. Kapadia, G.~Westheimer, and C.~D. Gilbert, ``Spatial distribution of
  contextual interactions in primary visual cortex and in visual perception,''
  {\em Journal of Neurophysiology}, vol.~84, no.~4, pp.~2048--2062, 2000.
\newblock doi: https://doi.org/10.1152/jn.2000.84.4.2048.

\bibitem{kapadia1995improvement}
M.~K. Kapadia, M.~Ito, C.~D. Gilbert, and G.~Westheimer, ``Improvement in
  visual sensitivity by changes in local context: Parallel studies in human
  observers and in v1 of alert monkeys,'' {\em Neuron}, vol.~15, no.~4,
  pp.~843--856, 1995.
\newblock doi: https://doi.org/10.1016/0896-6273(95)90175-2.

\bibitem{altmann2003perceptual}
C.~F. Altmann, H.~H. B{\"u}lthoff, and Z.~Kourtzi, ``Perceptual organization of
  local elements into global shapes in the human visual cortex,'' {\em Current
  Biology}, vol.~13, no.~4, pp.~342--349, 2003.
\newblock doi: https://doi.org/10.1016/S0960-9822(03)00052-6.

\bibitem{boykov2001interactive}
Y.~Y. Boykov and M.-P. Jolly, ``Interactive graph cuts for optimal boundary \&
  region segmentation of objects in nd images,'' in {\em 8th IEEE International
  Conference on Computer Vision}, pp.~105--112, 2001.
\newblock doi: https://doi.org/10.1109/ICCV.2001.937505.

\bibitem{boykov2004experimental}
Y.~Boykov and V.~Kolmogorov, ``An experimental comparison of min-cut/max-flow
  algorithms for energy minimization in vision,'' {\em IEEE Transactions on
  Pattern Analysis and Machine Intelligence}, vol.~26, no.~9, pp.~1124--1137,
  2004.
\newblock doi: https://doi.org/10.1109/TPAMI.2004.60.

\bibitem{field1993contour}
D.~J. Field, A.~Hayes, and R.~F. Hess, ``Contour integration by the human
  visual system: Evidence for a local “association field”,'' {\em Vision
  Research}, vol.~33, no.~2, pp.~173--193, 1993.
\newblock doi: https://doi.org/10.1016/0042-6989(93)90156-Q.

\bibitem{huyer1999global}
W.~Huyer and A.~Neumaier, ``Global optimization by multilevel coordinate
  search,'' {\em Journal of Global Optimization}, vol.~14, no.~4, pp.~331--355,
  1999.
\newblock doi: https://doi.org/10.1023/A:1008382309369.

\bibitem{wertheimer1938laws}
M.~Wertheimer, ``Laws of organization in perceptual forms,'' in {\em A Source
  Book of Gestalt Psychology} (W.~D. Ellis, ed.), pp.~71--88, London, U.K.:
  Routledge \& Kegan Paul, 1938.
\newblock doi: https://doi.org/10.1037/11496-005 (Original work published
  1923).

\bibitem{hough1962method}
P.~V. Hough, ``Method and means for recognizing complex patterns,'' dec 1962.
\newblock US Patent 3,069,654.

\bibitem{akinlar2011edlines}
C.~Akinlar and C.~Topal, ``Edlines: A real-time line segment detector with a
  false detection control,'' {\em Pattern Recognition Letters}, vol.~32,
  no.~13, pp.~1633--1642, 2011.
\newblock doi: https://doi.org/10.1016/j.patrec.2011.06.001.

\bibitem{akinlar2013edcircles}
C.~Akinlar and C.~Topal, ``Edcircles: A real-time circle detector with a false
  detection control,'' {\em Pattern Recognition}, vol.~46, no.~3, pp.~725--740,
  2013.
\newblock doi: https://doi.org/10.1016/j.patcog.2012.09.020.

\bibitem{geisler2001edge}
W.~S. Geisler, J.~S. Perry, B.~Super, and D.~Gallogly, ``Edge co-occurrence in
  natural images predicts contour grouping performance,'' {\em Vision
  Research}, vol.~41, no.~6, pp.~711--724, 2001.
\newblock doi: https://doi.org/10.1016/S0042-6989(00)00277-7.

\bibitem{cao2019application}
Y.-J. Cao, C.~Lin, Y.-J. Pan, and H.-J. Zhao, ``Application of the
  center--surround mechanism to contour detection,'' {\em Multimedia Tools and
  Applications}, vol.~78, no.~17, pp.~25121--25141, 2019.
\newblock doi: https://doi.org/10.1007/s11042-019-7722-1.

\bibitem{parent1989trace}
P.~Parent and S.~W. Zucker, ``Trace inference, curvature consistency, and curve
  detection,'' {\em IEEE Transactions on Pattern Analysis and Machine
  Intelligence}, vol.~11, no.~8, pp.~823--839, 1989.
\newblock doi: https://doi.org/10.1109/34.31445.

\bibitem{hummel1983foundations}
R.~A. Hummel and S.~W. Zucker, ``On the foundations of relaxation labeling
  processes,'' {\em IEEE Transactions on Pattern Analysis and Machine
  Intelligence}, vol.~PAMI-5, no.~3, pp.~267--287, 1983.
\newblock doi: https://doi.org/10.1109/TPAMI.1983.4767390.

\bibitem{farag1995edge}
A.~A. Farag and E.~J. Delp, ``Edge linking by sequential search,'' {\em Pattern
  Recognition}, vol.~28, no.~5, pp.~611--633, 1995.
\newblock doi: https://doi.org/10.1016/0031-3203(94)00131-5.

\bibitem{ji2013sequential}
X.~Ji, X.~Zhang, and L.~Zhang, ``Sequential edge linking method for
  segmentation of remotely sensed imagery based on heuristic search,'' in {\em
  21st International Conference on Geoinformatics}, 2013.
\newblock doi: https://doi.org/10.1109/Geoinformatics.2013.6626164.

\bibitem{lu2008edge}
D.-S. Lu and C.-C. Chen, ``Edge detection improvement by ant colony
  optimization,'' {\em Pattern Recognition Letters}, vol.~29, no.~4,
  pp.~416--425, 2008.
\newblock doi: https://doi.org/10.1016/j.patrec.2007.10.021.

\bibitem{topal2012edge}
C.~Topal and C.~Akinlar, ``Edge drawing: a combined real-time edge and segment
  detector,'' {\em Journal of Visual Communication and Image Representation},
  vol.~23, no.~6, pp.~862--872, 2012.
\newblock doi: https://doi.org/10.1016/j.jvcir.2012.05.004.

\bibitem{akinlar2015cannysr}
C.~Akinlar and E.~Chome, ``Cannysr: Using smart routing of edge drawing to
  convert canny binary edge maps to edge segments,'' in {\em International
  Symposium on Innovations in Intelligent SysTems and Applications}, 2015.
\newblock doi: https://doi.org/10.1109/INISTA.2015.7276784.

\bibitem{akinlar2016pel}
C.~Akinlar and E.~Chome, ``Pel: A predictive edge linking algorithm,'' {\em
  Journal of Visual Communication and Image Representation}, vol.~36,
  pp.~159--171, 2016.
\newblock doi: https://doi.org/10.1016/j.jvcir.2016.01.017.

\bibitem{seo2019subpixel}
S.~Seo, ``Subpixel line localization with normalized sums of gradients and
  location linking with straightness and omni-directionality,'' {\em IEEE
  Access}, vol.~7, pp.~180155--180167, 2019.
\newblock doi: https://doi.org/10.1109/ACCESS.2019.2959320.

\bibitem{ernst2012optimality}
U.~A. Ernst, S.~Mandon, N.~Schinkel-Bielefeld, S.~D. Neitzel, A.~K. Kreiter,
  and K.~R. Pawelzik, ``Optimality of human contour integration,'' {\em PLoS
  Computational Biology}, vol.~8, no.~5, 2012.
\newblock doi: https://doi.org/10.1371/journal.pcbi.1002520.

\bibitem{koller2009probabilistic}
D.~Koller and N.~Friedman, {\em Probabilistic graphical models: Principles and
  techniques}.
\newblock Cambridge, MA, U.S.: MIT press, 2009.
\newblock doi: https://doi.org/10.1017/S0269888910000275.

\bibitem{jegelka2011submodularity}
S.~Jegelka and J.~Bilmes, ``Submodularity beyond submodular energies: Coupling
  edges in graph cuts,'' in {\em Computer Vision and Pattern Recognition},
  pp.~1897--1904, 2011.
\newblock doi: https://doi.org/10.1109/CVPR.2011.5995589.

\bibitem{kolmogorov2004energy}
V.~Kolmogorov and R.~Zabin, ``What energy functions can be minimized via graph
  cuts?,'' {\em IEEE Transactions on Pattern Analysis and Machine
  Intelligence}, vol.~26, no.~2, pp.~147--159, 2004.
\newblock doi: https://doi.org/10.1109/TPAMI.2004.1262177.

\end{thebibliography}
 \bibliographystyle{ieeetr}

\end{document}